%% file: c2a_paper.tex
\documentclass{vgtc}                

\usepackage{graphicx}
\usepackage{subfigure}
\usepackage{caption}
\usepackage{caption}
\usepackage{morefloats}
\usepackage{enumitem}
\usepackage{multirow}
\usepackage{array}
\usepackage{dblfloatfix}

\usepackage{microtype}                 
\PassOptionsToPackage{warn}{textcomp}  
\usepackage{textcomp}                  
\usepackage{mathptmx}                  
\usepackage{times}                     
\usepackage{cite}
\usepackage{color,soul}
\def\href#1#2{#1 #2}
\soulregister\cite7
\soulregister\ref7

\makeatletter
 \def\SOUL@hlpreamble{%
 \setul{}{2.4ex}
 \let\SOUL@stcolor\SOUL@hlcolor
 \SOUL@stpreamble
 }
\makeatother

\newcolumntype{P}[1]{>{\centering\arraybackslash}p{#1}}

\onlineid{0}

\vgtccategory{Research}

\title{C$^{2}$A: Crowd Consensus Analytics for Virtual Colonoscopy}

\author{Ji~Hwan~Park\thanks{e-mail:jihwpark@cs.stonybrook.edu},~Saad~Nadeem\thanks{e-mail:sanadeem@cs.stonybrook.edu},~Seyedkoosha~Mirhosseini\thanks{e-mail:smirhosseini@cs.stonybrook.edu},~and~Arie~Kaufman\thanks{e-mail:ari@cs.stonybrook.edu}}
\affiliation{\scriptsize Department of Computer Science, Stony Brook University}

\authorfooter{
E-mail: \{jihwpark,sanadeem,smirhosseini,ari\}@cs.stonybrook.edu}

\abstract{
We present a medical crowdsourcing visual analytics platform called C{$^2$}A to visualize, classify and filter crowdsourced clinical data. More specifically, C{$^2$}A is used to build consensus on a clinical diagnosis by visualizing crowd responses and filtering out anomalous activity. Crowdsourcing medical applications have recently shown promise where the non-expert users (the crowd) were able to achieve accuracy similar to the medical experts. This has the potential to reduce interpretation/reading time and possibly improve accuracy by building a consensus on the findings beforehand and letting the medical experts make the final diagnosis. In this paper, we focus on a virtual colonoscopy (VC) application with the clinical technicians as our target users, and the radiologists acting as consultants and classifying segments as benign or malignant. In particular, C{$^2$}A is used to analyze and explore crowd responses on video segments, created from fly-throughs in the virtual colon. C{$^2$}A provides several interactive visualization components to build crowd consensus on video segments, to detect anomalies in the crowd data and in the VC video segments, and finally, to improve the non-expert user's work quality and performance by A/B testing for the optimal crowdsourcing platform and application-specific parameters. Case studies and domain experts feedback demonstrate the effectiveness of our framework in improving workers' output quality, the potential to reduce the radiologists' interpretation time, and hence, the potential to improve the traditional clinical workflow by marking the majority of the video segments as benign based on the crowd consensus.
} 


\CCScatlist{ 
\CCScat{Human-centered computing}{Visualization}{Visualization
application domains}{Visual analytics}
}

   \teaser{
   \centering
\vspace{-4mm}
   \includegraphics[width=16cm]{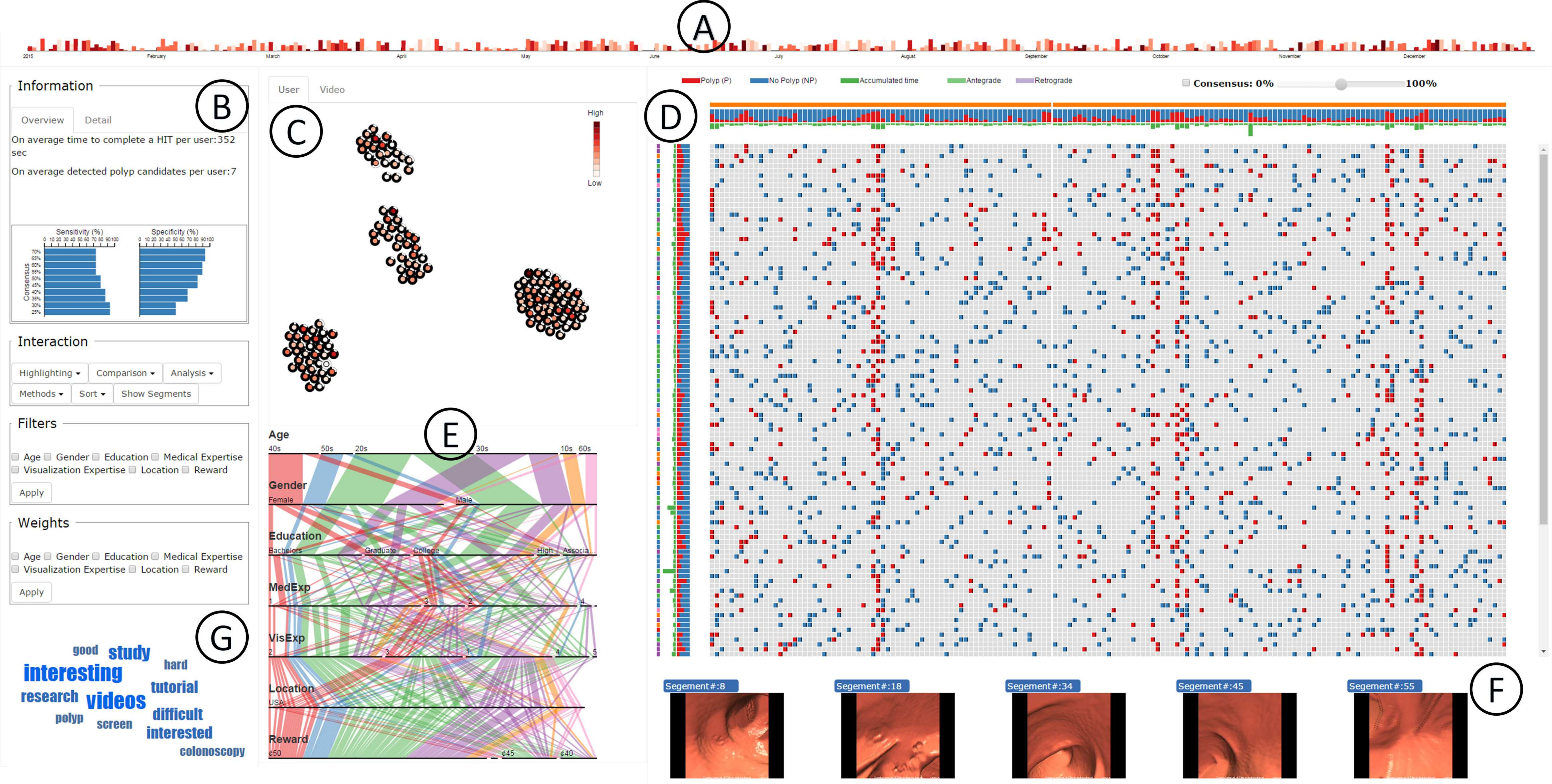}
\vspace{-2mm}
   \caption{The C${^2}$A system includes (A) Timeline Filtering View for selecting datasets within a specified time interval, (B) Aggregated Textual Information for displaying a textual summary of the crowd statistics and application specific performance, (C) Similarity View for displaying the overlap and Euclidean distance metric of the crowd demographics and video segment statistics, (D) Consensus Map for displaying the crowd consensus on polyp and polyp-free (benign) video segments along with aggregated crowd accuracy and timing, (E) Crowd View for displaying crowd demographics and rewards, (F) Video Segments View for displaying selected video segments, and (G) Word Cloud displaying keywords from user comments.}
\label{fig:teaser}
  }

\vgtcinsertpkg

\begin{document}



\input{sec-introduction}
\input{sec-related-work}
\input{sec-cvc}
\input{sec-design}
\input{sec-usecase}
\input{sec-discussion}
\input{sec-conclusion}

\acknowledgments{
We are grateful to Drs. Kevin Baker and Matthew Barish, Radiology Department, Stony Brook University Hospital, for their critical feedback and guidance. This research has been partially supported by the NSF grants CNS0959979, IIP1069147, and CNS1302246.}


\end{document}

%% file: sec-introduction.tex
\firstsection{Introduction}
\maketitle
There is a recent surge in non-invasive imaging techniques to screen patients for abnormalities, including cancers. However, this has also led to a significant increase in the radiologists' interpretation time of these non-invasive imaging cases, such as virtual colonoscopy, lung nodule detection, and breast mammography.

\begin{figure}[t]
\vspace{-2mm}
\centering
\begin{subfigure}[]{
\includegraphics[width=0.22\textwidth]{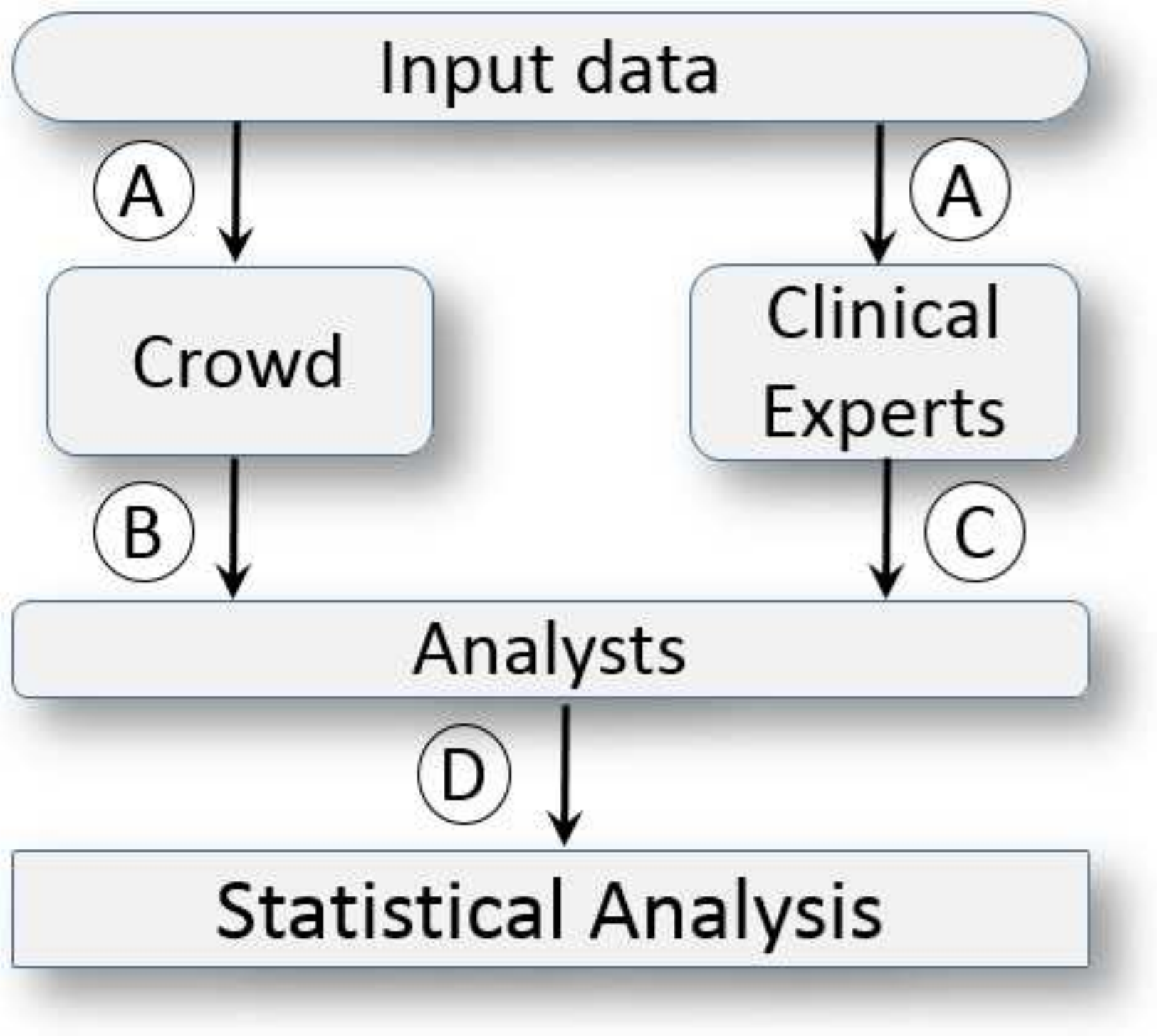}}
\end{subfigure}
\begin{subfigure}[]{
\includegraphics[width=0.22\textwidth]{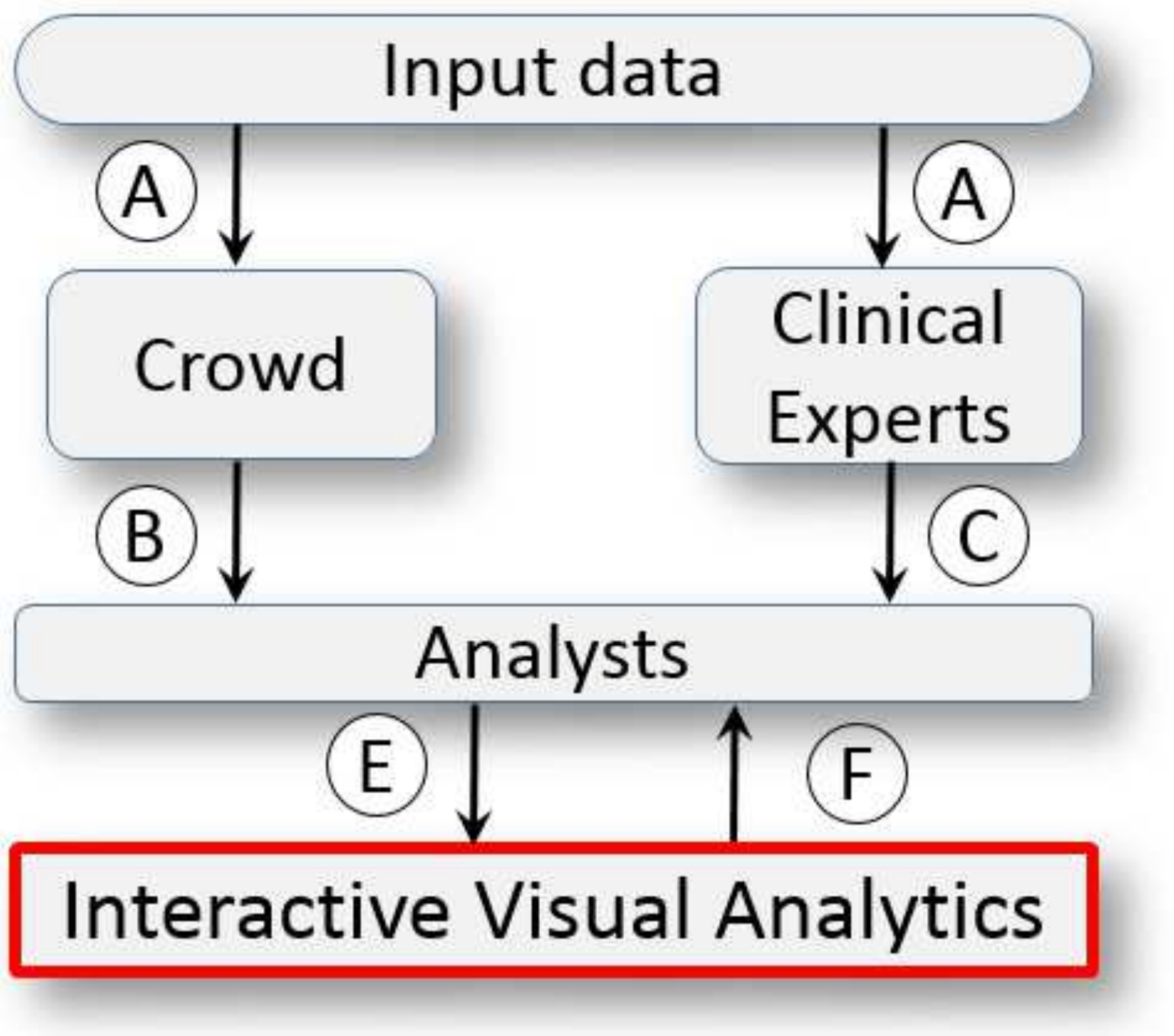}}
\end{subfigure}
\vspace{-5mm}
\caption{(a) An existing workflow for medical crowdsourcing~\cite{Mortensen:2015} and (b) our workflow. First, input data (colon video segments, in our case) are created and (A) presented to the crowd (non-expert users) and the clinical experts (radiologists in our case). The crowd (B) and the clinical experts (C) conduct micro-tasks (identifying each video segment as a polyp/polyp-free video segment, in our case), simultaneously. Once the results from the crowd and the radiologists are obtained, in the existing workflow~\cite{Mortensen:2015}, analysts perform statistical analysis based on the selected criteria (D). However, in our workflow, analysts (clinical technicians) can interactively explore, analyze, and compare the results from the crowd with the ground truth from the medical experts using C{$^2$}A (E) and (F).}
\label{fig:pipeline}
\vspace{-2mm}
\end{figure}

Computer-aided polyp detection (CAD) techniques have been introduced to aid radiologists~\cite{hong:2006:tvcg,zhao:2006lines} but the miss rate for these techniques is high due to the complexity and variability in the shape and characteristics of these precursor cancerous lesions~\cite{Summers:2008}. The crowd of non-expert users, however, has fared much better in comparison to these CAD techniques in reducing the false positives (benign structures interpreted as cancerous lesions) and false negatives (missed cancerous lesions), and achieving specificity and sensitivity comparable to the expert radiologists in specific biomedical applications with some initial training~\cite{Nguyen:2012,McKenna:2012}.
In this paper, we extend these findings to create C${^2}$A, a visual analytics platform (Fig.~\ref{fig:teaser}) for clinical technicians or analysts to build a consensus on crowd findings, to detect anomalies from the crowd, and to improve the quality of a worker's output by doing A/B testing for platform and application-specific parameters. The clinical technicians in our context are the intermediaries between the crowd and the radiologists who conduct the medical crowdsourcing studies or the analysts running the crowdsourcing application. This intermediary can also be a hospital technician responsible for preparing the medical data for the radiologist's examination. We will refer to this intermediary as a clinical technician in this paper. Fig.~\ref{fig:pipeline} shows our workflow.

\begin{figure}[t]
\vspace{-1mm}
\centering
\begin{subfigure}[]{
\includegraphics[width=0.17\textwidth]{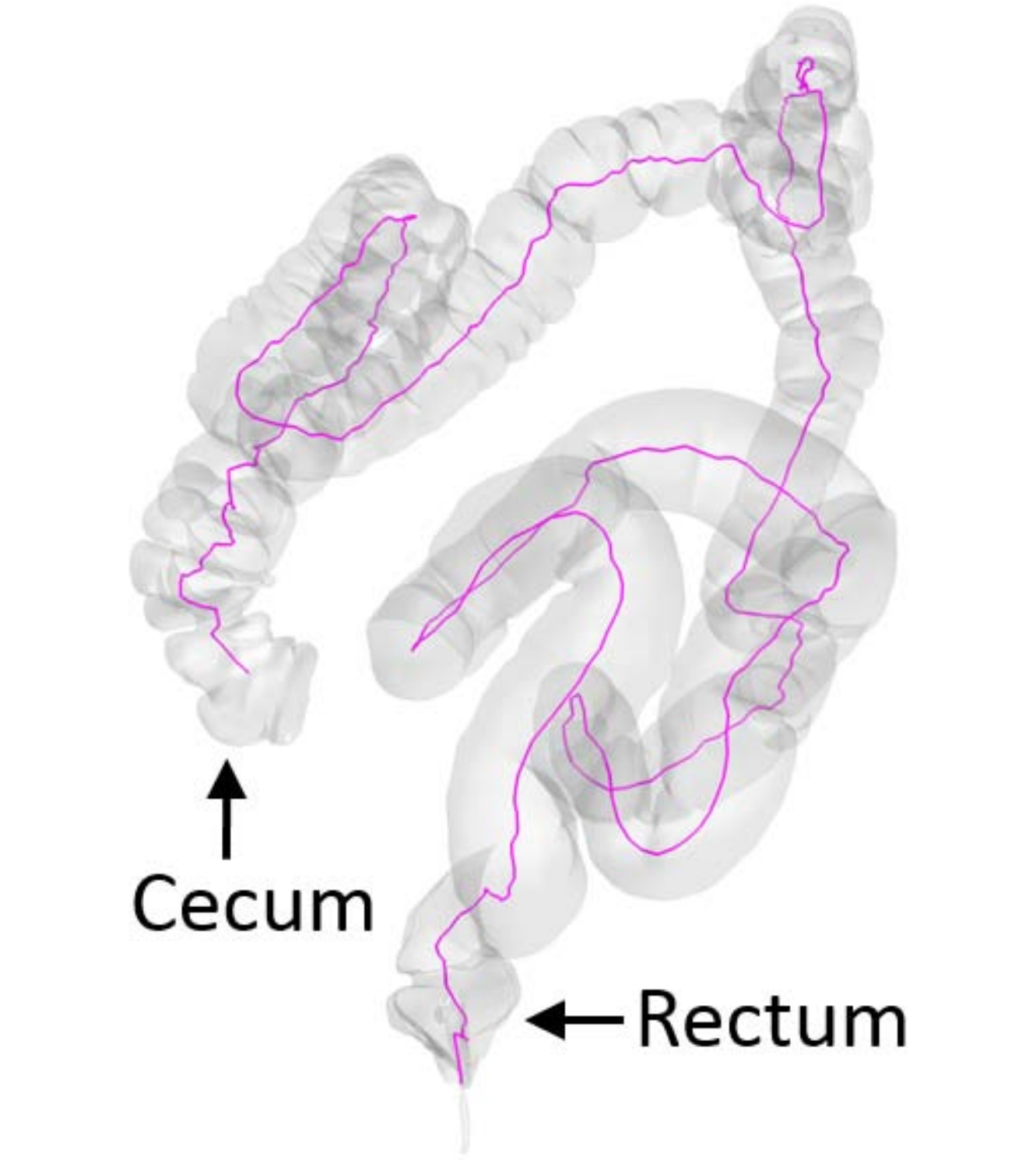}}
\end{subfigure}
\begin{subfigure}[]{
\includegraphics[width=0.16\textwidth]{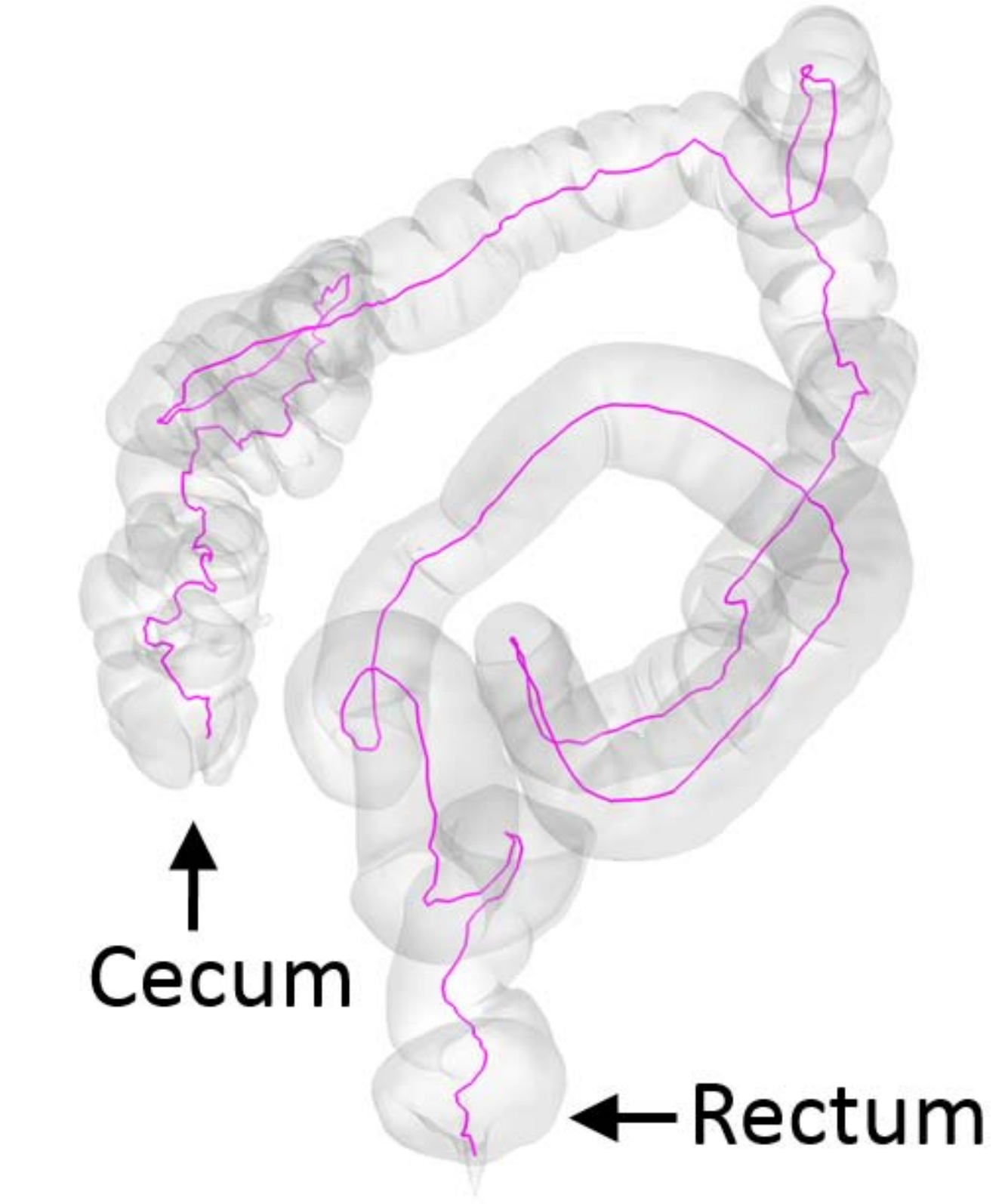}}
\end{subfigure}
\vspace{-5mm}
\caption{The (a) supine and (b) prone colons, and the corresponding computed centerlines (shown in pink) for creating the virtual fly-throughs in VC, from rectum to cecum (antegrade direction) and cecum to rectum (retrograde direction).}
\label{fig:colon}
\vspace{-3mm}
\end{figure}

VC is a non-invasive screening procedure for colorectal cancer in which a 3D colon is reconstructed from a CT scan of the patient's abdomen. The radiologists navigate from the rectum to the cecum (antegrade) and back
(retrograde) through a reconstructed 3D colon (Fig.~\ref{fig:colon}) while inspecting the colon wall for polyps, the precursors of colorectal cancer~\cite{hong:1997:siggraph}.
We conducted interviews with five VC-trained radiologists; one of whom is a world renowned expert on VC. According to these radiologists, VC interpretation can take 30 minutes on average.
This is a significant portion of the radiologists' time considering that they have to read numerous cases a day and hence time is a bottleneck for them. With the recent assignment of a reimbursement code for VC and stronger compliance of patients (above age 50) for VC, the number of VC cases are bound to increase.
Hence, a more reliable assistance in the form of the crowd can significantly reduce the interpretation time and allow radiologists to read more cases, thus increasing their productivity. Considering these benefits, the radiologists interviewed for this study were willing to bear the cost of prior crowd interpretation by letting go of a meager fraction of their commission, for convenience and for a stronger corroboration of their final diagnosis. Hospitals can benefit from the crowd assistance as well. Hospitals, for instance, often require multiple radiologists to read the same cases to avoid any unnecessary lawsuits for misdiagnosis. In this context, a minimally trained group of non-expert users can provide a cheaper and effective alternative for the hospitals. Finally, the radiologists interviewed were of the opinion that any additional costs, if incurred, will trickle down to the insurance companies but not to the patients. In the future, we will engage all the stakeholders (hospitals, radiologists, and patients) to get a more thorough perspective on this issue.

We recently conducted a preliminary study~\cite{Park:2016} to leverage the crowd to detect polyp and polyp-free (benign) segments for a given VC dataset with sensitivity and specificity comparable to the radiologists. More specifically, we created centerline fly-through videos in the antegrade and retrograde directions for patient VC datasets in supine (facing up) and prone (facing down) orientations (see Fig.~\ref{fig:colon}). We then divided these videos into equi-timed segments with some overlap to make these more conducive to crowdsourcing. These short video segments were presented to the crowd for detecting polyps and polyp-free video segments after some basic training. We showed that  in most cases, the majority of the video segments can be discarded as polyp-free based on the crowd consensus, reducing the overall interpretation time. This detection of polyp and polyp-free segments, in our view, holds the highest potential for reducing the radiologists' interpretation time by letting them focus on the segments containing polyps.

In this work, we leverage the promising results for crowdsourcing VC to create a visual analytics platform for clinical technicians to interpret the crowd data. The goals of this platform underline A/B testing for a crowdsourcing platform and application-specific parameters, anomaly detection in crowd and video segments, and creating a consensus on the video segments for eventual reading by radiologists.

More data results in more noise. Clinical technicians running the crowdsourcing medical studies can leverage C{$^2$}A to manage this noise and apply appropriate filters and consensus thresholds to find the optimal specificity and sensitivity values that are comparable to the radiologists. Hence C{$^2$}A will be used recurrently to reach the best consensus across different datasets. Since unreliable users can skew the final results, clinical technicians can find these users via different clustering and aggregation views provided in C{$^2$}A. Moreover, the changing requirements (for example, no embedded javascript) and incentive models (rewards and bonuses) on the prevalent crowdsourcing platforms (e.g., Amazon Mechanical Turk (MTurk)~\cite{mturk} and CrowdFlower~\cite{crowdflower}) also mean that the clinical technicians will have to consistently test for these platform-specific parameters using the different views provided in C{$^2$}A. Hence, C{$^2$}A has the potential to become a critical component in longitudinal crowdsourcing medical studies and applications. Our contributions are:
\begin{itemize}[noitemsep,topsep=0pt]
  \item We develop an interactive visual system for clinical technicians to assist radiologists in finding optimal parameters to obtain the best results from a crowdsourced VC application, detecting anomalies in the crowd and video segments, and building a consensus on polyp and polyp-free video segments, with sensitivity and specificity comparable to expert radiologists.
  \item We show the effectiveness of our platform by applying it to real VC datasets and crowd results.
\end{itemize} 

%% file: sec-related-work.tex
\section{Related Work}
\label{sec:previous}

Crowdsourcing approaches are popular in various domains and there have been attempts to leverage non-expert workers in medicine.
Maier-Hein et al.~\cite{Maier-Hein2014} have used the crowd to annotate endoscopic video images, and showed that annotations from the crowd are similar to those of medical experts. Another study has shown that the crowd has comparable accuracy to medical experts in segmenting objects in various biomedical images~\cite{Gurari:2015}.
Miltry et al.~\cite{Mitry:2013} have shown the effectiveness of using crowdsourcing in classifying retinal fundus photograph images.
Nguyen et al.~\cite{Nguyen:2015} have used both crowd workers and experts to screen biomedical citation, and their method outperformed screening by crowds only.
Two studies harnessed crowdsourcing to identify false positive polyps from polyp candidates in VC~\cite{Nguyen:2012,McKenna:2012}.
In the first work, participants/workers were shown images of 3D reconstructions of the polyp candidates, whereas in the follow-up work they were shown the videos of polyp candidates for better depth perception. In the first work, the results showed that there was no significant difference between the crowd and CAD and the crowd was in fact better than CAD on easy polyp candidates. In the second work, the depth perception in the videos improved the detection rate of true polyps and the crowd was better than CAD on easy and moderate polyp candidates. Moreover, the top workers performed at an accuracy similar to that of an expert radiologist. These papers have highlighted superior performance and accuracy of non-expert users in comparison to CAD and the potential of the crowd to be used as primary detectors for polyps rather than just being used for false positive reduction.
However, in order to utilize non-expert workers for clinical tasks, we should control workers' quality to maximize their performance because low quality results from these workers can make it difficult to build consensus on the polyp and polyp-free segments and can hamper the ultimate goal of reducing radiologists' interpretation time.

There are studies on improving workers' performance.
Willett et al.~\cite{Willett:2012} have provided principles to improve the quality of workers' output for social data analysis.
Gadiraju et al.~\cite{Gadiraju:2014} have classified crowdsourcing tasks into six categories based on their goal.
They also identified characteristics of each category in terms of effort, task completion rate, and rewards satisfaction, and found that workers selected tasks based on rewards, interest, and time to complete.

Visual analytics can be powerful for crowd-assisted VC because identifying anomaly workers and reducing false negative polyps are crucial in VC.
There is no previous work on crowd consensus analytics for biomedical applications. Instead, we describe several approaches related to components of our platform.
Willett et al.~\cite{willett:2013} have presented a framework to cluster and interpret results from crowd workers by other crowd workers.
CrowdScape~\cite{Rzeszotarski:2012} is a system to visualize workers' behavior and their output to evaluate the quality of their answers, allows users to interactively explore these two features, and enables users to classify workers.
Mimic~\cite{Breslav:2014} is a system to help interaction designers understand the relationship between workers' output and their behavior.
Unlike CrowdScape, it focuses on micro interactions.
CrowdWeaver~\cite{Kittur:2012} is a system to create and organize tasks for the crowd, where each task is saved and can be reused.
Additionally, it allows users to manage the quality of tasks.
DemographicsVis~\cite{Dou:2015} is a framework to analyze relationships between demographic information and user-created data.

To the best of our knowledge, there are no crowd consensus analytics platforms for crowdsourced VC. In effect, our C{$^2$}A platform is a critical component for clinical technicians to visualize different relationships between parameters and users' performance and to explore and analyze crowd results for identification of polyp and polyp-free segments and thus, potentially reduce the radiologists' interpretation time by letting them focus on segments with polyps. 

%% file: sec-cvc.tex
\section{Crowd Consensus Analytics for VC}
\label{sec:cvc}

In our work, we present a crowd consensus analytics platform for a VC application. In this section, we explain the data generation and input for the crowdsourcing platform and enumerate the tasks that need to be analyzed for the eventual crowd consensus on polyp and polyp-free segments.

\subsection{Data Generation and Collection} 
\label{sec:data}

\subsubsection{Endoluminal Video Segments}
In clinical VC systems, the standard view is a virtual fly-through of a rendered colon model which mimics the general appearance of an optical colonoscopy. This rendering presents an accurate and understandable reconstruction of an endoluminal view inside the colon. As a first pass in screening, the radiologist typically flies through the colon using this endoluminal view, examining the reconstructed mucosa for any polyps, which appear as protrusions on the wall.

Since this endoluminal view is the standard view in VC and the view most easily understood for non-experts, we render such views as the type of imagery to present to the crowd workers. To create this view, a virtual camera is placed along a path, and raycasting from the camera position through the CT volume results in the final generated image. Typically, a centerline is calculated through the colon lumen and is used as the flight path, yielding good general coverage. However, due to the haustral folds and bends of the colon, some portions are missed~\cite{hong:2007:spie}, necessitating both antegrade and retrograde fly-throughs of each virtual colon model.

For this research, we have used the commercial FDA approved Viatronix V3D-Colon VC system~\cite{viatronix} to generate the endoluminal video segments. After the fly-throughs were established and videos were captured, the traversal through the colon was divided into approximately equi-time segments. These segments have slight overlap to ensure that a polyp which might be located on a video segment's boundary is not missed.

When generating the fly-through videos, it is possible to adjust some parameters that are used for the rendering. One parameter is the flight speed, allowing users to fly slower or faster based on their experience. While a faster speed requires less reading time, it can also lead to areas being overlooked as they appear only briefly and a dedicated user might have to go back and forth to detect a polyp. A second parameter is the field of view (FOV); a wide angle fisheye view with a 120 degree viewing angle can be enabled to allow for greater coverage of the colon wall. While more of the wall becomes visible in a single fly-through, this option introduces significant distortion, which might cause confusion for the viewer.

\subsubsection{Video Data and Ground Truth}
The video data was generated from 4 patients, with a total of 16 complete fly-through videos being captured; 2 patient datasets were used to generate 163 video segments with 90 degree FOV and 50 frames/second, and the other 2 were used to generate 136 video segments with 120 degree FOV and 100 frames/second 136 video segments. Both these video segment datasets include antegrade and retrograde navigation in both the supine (patient facing up) and prone (patient facing down) scanned VC datasets.

The colon CT data which was used for this study had previously undergone VC examination, with each patient having been found to have at least one polyp which was later clinically confirmed with biopsy. Since these polyps had been previously located, we were able to note at what time during the full video fly-throughs each polyp would come into view and exit the view. These ground truth annotations were used when splitting the videos to identify which segment contained a polyp.

\subsubsection{User Interface}

To present our study to the non-expert crowd workers, we make use of the MTurk platform, which has recently become popular as a way of obtaining reliable crowd workers at modest cost while allowing us to reach a large and diverse population of users~\cite{paolacci:2010}. When a user first selects this task, s/he is provided with brief directions about the purpose of the system as well as short tutorial videos. Each of these videos contains a polyp, which is annotated in order to illustrate to the workers what they should be looking for.

After viewing the brief tutorial, the user is sequentially provided with twenty video segments. For each video, there is an option for Yes (a polyp candidate is present in the current video) or No (there is no polyp candidate present in the current video). After selecting one of the options, the user submits the result and continues directly to the next video. The videos start playing automatically when the page loads, and the user can pause the video, replay the video, and drag the time slider as desired to view specific frames of interest. Once the user has clicked Submit, the user cannot go back to a previous video. After the user has inspected all twenty videos, comments are requested from the user, which we use to refine our system.

The presentation of the videos was randomized between participants. Each user is allowed to perform the task only once. In the future, we will allow reliable users with high accuracy to perform this task multiple times and provide them with additional incentives.

\subsection{Task analysis} 
\label{sec:task}
We worked with the radiologists to identify the tasks that can assist in effective consensus building for VC. In particular, we extend our discussion and insights from our recent findings~\cite{Park:2016} to create a pool of tasks that can be used to improve the quality of workers among the crowd, to detect the anomalies in the crowd and the video segments, and to build a stronger consensus on polyp and polyp-free segments.

\paragraph{T1. Effects of parameters on user performance and accuracy}
Through our discussion with the expert radiologists and previous crowdsourcing research, several demographic, incentive, and application-specific parameters have been found to be important for building an appropriate consensus on the annotation tasks. For example, \emph{How do the rewards and bonuses affect the user performance and accuracy? How does the location affect the user performance and accuracy? Does the length of the video segments affect the user attention span and hence the performance and accuracy? What value of field-of-view can create enough distortion to reduce the user performance and accuracy? What speed of fly-through causes the user accuracy and performance to drop?}

\paragraph{T2. Anomaly detection}
Through our recent work, we found that the anomalous users and video segments can skew the consensus considerably. In effect, we try to answer the following questions to narrow the search space for anomalous users and video segments: \emph{Are there users who complete the task in less than the allocated time for the whole task? Are there video segments where there is no consensus? Are there users who randomly click yes or no for answers regardless of time?}

\paragraph{T3. Crowd consensus on the video segments}
Through our discussions with expert radiologists, we found that finding an appropriate threshold of users to build a strong consensus is important for monetary reasons. For example, \emph{How many users are needed to reach an ideal tradeoff for sensitivity and specificity, comparable to expert radiologists? What will be the associated cost of the users to build a strong consensus?}


%% file: sec-design.tex
\begin{table}[tb]
\small
\renewcommand{\arraystretch}{1.1} 
\caption{The consensus map can visualize two types of information with the corresponding elements and the aggregated information.}
\label{Tab:typeofdata}
\vspace{-3mm}
\centering
\begin{tabular}{P{0.06\textwidth}|P{0.07\textwidth}|p{0.3\textwidth}}

  \hline
  Information \newline Type &  Elements		&  \multicolumn{1}{c}{Aggregated Information} 	 \\
  \hline   \hline
User\newline  Response & Polyp, Polyp-free &
\textbf{User aggregated information}: Total number of polyp and polyp-free video segments detected\newline
\textbf{Aggregated user time to complete a task} (20 video segments) normalized using time across all users\newline									
\textbf{Video segment aggregated information}: Total number of users marking this video segment as polyp or polyp-free \newline
\textbf{Aggregated time of users to complete a video segment} normalized using time across all video segments\\
 \hline
Statistics & Correct, False positive, False negative &
\textbf{User aggregated information}: Total number of a user's correct, false positive, and false negative responses\newline
\textbf{Aggregated user time to complete a task} (20 video segments) normalized using time across all users\newline
\textbf{Video segment aggregated information}: Total number of users marking this video segment as correct, false positive, and false negative\newline
\textbf{Aggregated time of users to complete a video segment} normalized using time across all video segments\\
  \hline
\end{tabular}
\vspace{-3mm}
\end{table}

\begin{figure*}[t]
\centering
\includegraphics[width=0.98\textwidth]{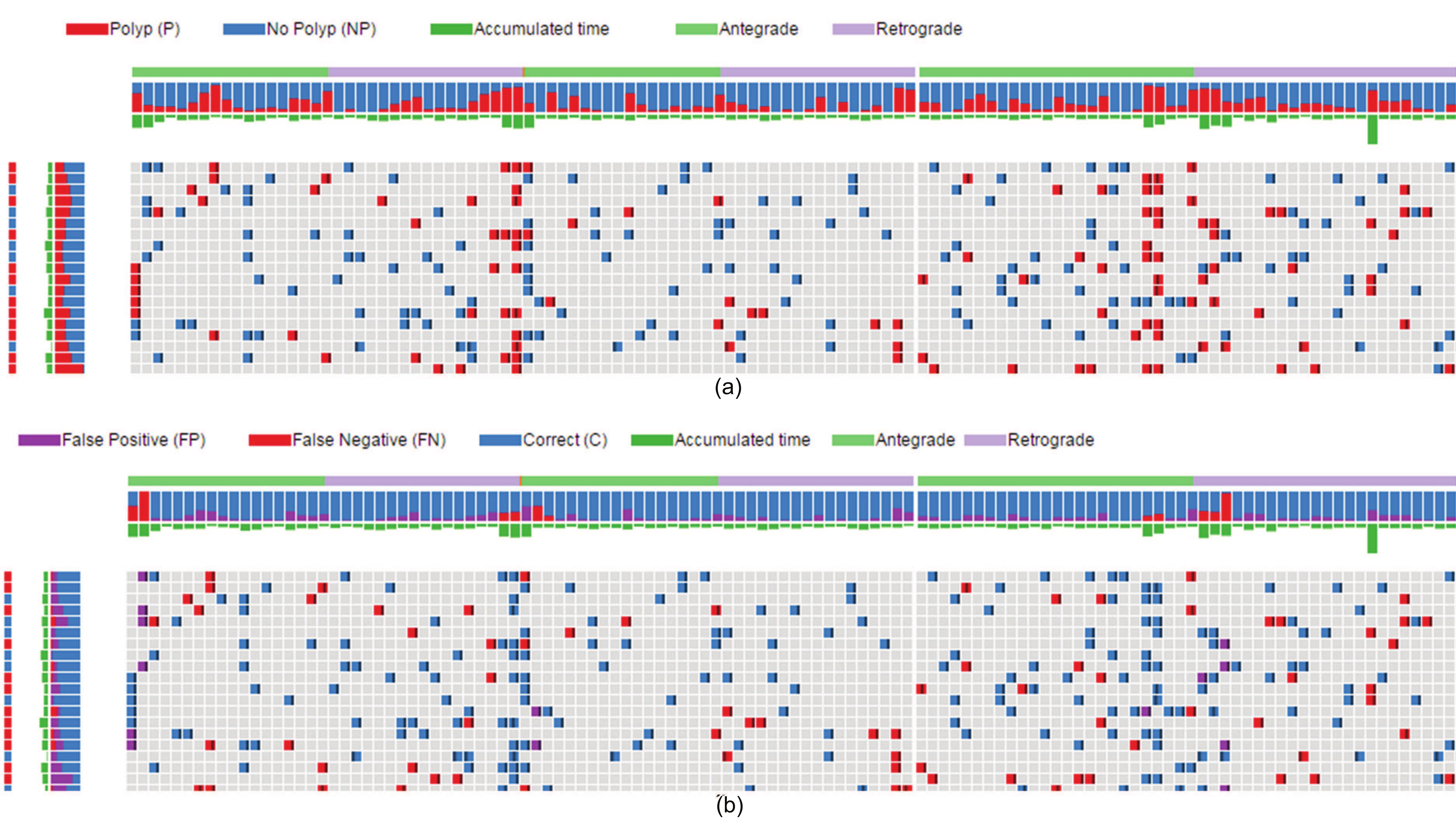}
\vspace{-3mm}
\caption{Two types of information in our consensus map view: (a) user response, and (b) statistics.}
\label{fig:consensusViews}
\vspace{-3mm}
\end{figure*}

\begin{figure}[htb!]
\centering
\includegraphics[width=0.48\textwidth]{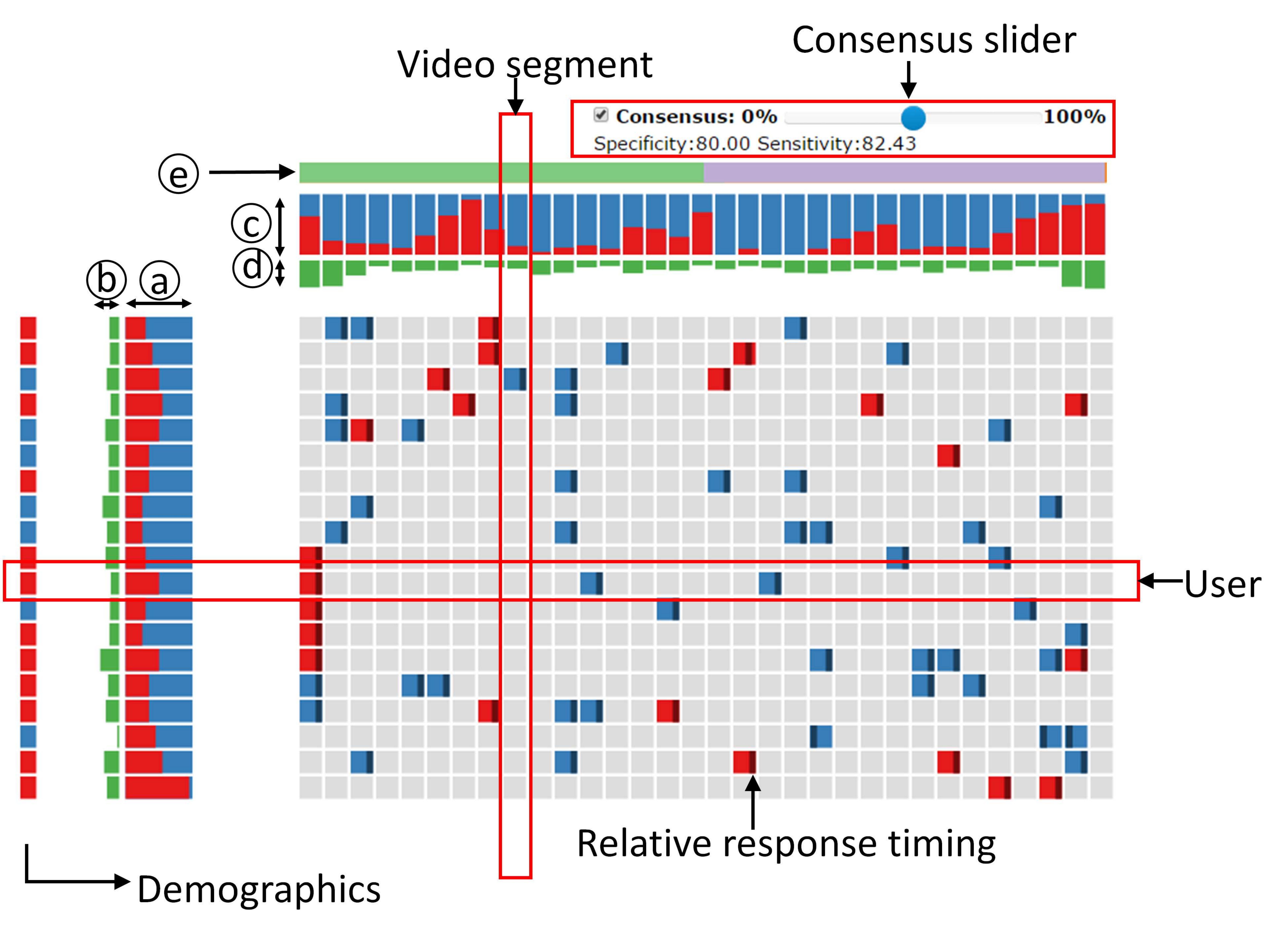}
\caption{Each cell of the matrix visualization in the consensus map displays the registered user response (if any), and the corresponding row and column represents a user and a video segment, respectively. We place a consensus slider on the top right corner of the map for selecting an appropriate threshold to obtain a consensus. On the left and top sides of the map, we also visualize, as enumerated in Table~\ref{Tab:typeofdata}, (a) User aggregated information, (b) Aggregated user time to complete a task (20 video segments) normalized using time across all users, (c) Video segment aggregated information, (d) Aggregated time of users to complete a video segment normalized using time across all video segments, and (e) the direction of fly-through, such as antegrade and retrograde. }
\label{fig:consensusmap}
\vspace{-5mm}
\end{figure}

\begin{figure*}[t]
\centering
\includegraphics[width=0.9\textwidth]{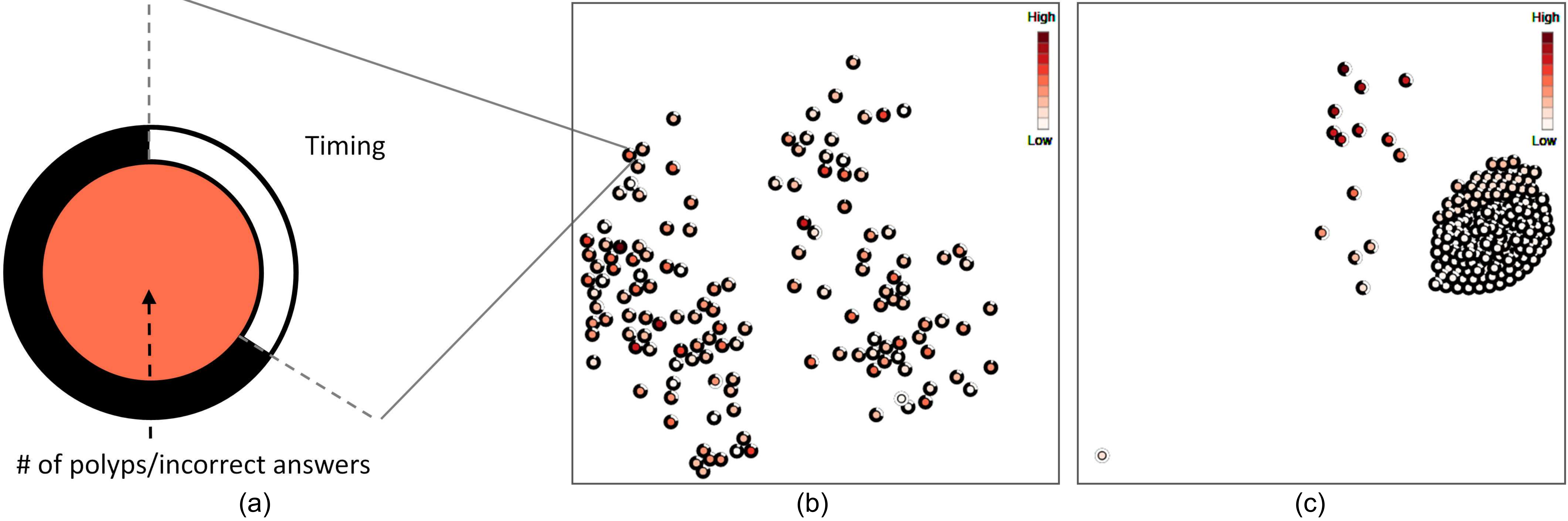}
\vspace{-4mm}
\caption{Illustration of our similarity view: (a) a glyph for each user and video segment used in our similarity view, (b) a similarity view based on users' demographics and rewards, and (c) a similarity view based on users' accuracy and timing on each video segment. In (a), the lightness of a circle indicates the number of polyps, respectively, and the length of an arc on a circle represent aggregated time to complete the task/video segment. }
\label{fig:similarityView}
\vspace{-3mm}
\end{figure*}

\section{Design}
\label{sec:design}

In order to effectively perform the tasks, we provide several linked views. When clinical technicians select a dataset in a timeline filtering view, all information related to the dataset is displayed in the main view, including the consensus map, the similarity view, the crowd view, aggregated textual information, and the word cloud.

\subsection{Consensus Map}
In order to visualize detailed information related to users' answers, we provide a consensus map to understand the relationship between users' answers and video segments (T1,T2,T3), as shown in Fig.~\ref{fig:teaser}(D). The consensus map can order users by time, by number of polyps, by accuracy, and by the number of false negatives.
The rows and columns in the consensus map denote users and video segments, respectively. Each cell of the consensus map can represent two types of information, namely about user response or about statistics, as shown in Table~\ref{Tab:typeofdata}. The information about user response is represented using two elements, polyp and polyp-free, as shown in Fig.~\ref{fig:consensusViews}(a). Similarly, the information regarding statistics is represented using three elements, namely correct, false positive, and false negative, as shown in Fig.~\ref{fig:consensusViews}(b). The aggregated user information on the left and the aggregated video segment information at the top in the consensus map, similarly, represent the two types of user response and statistics information (as depicted in Table~\ref{Tab:typeofdata} and Fig.~\ref{fig:consensusViews}).
We place a bar inside each cell to represent a relative response time to complete a video segment compared to other users who view the same video segment and the color of each bar is the same as the corresponding element, but darker.
In order to distinguish between all elements and aggregated information effectively, colors were selected from the ColorBrewer color scheme~\cite{Harrower:2003} for each aspect, and the key is located at the top of the map.
When we visualize aggregated response time, we normalize the time based on the longest time taken for the user response. On the left side in the consensus map view, we have also included a demographic label for each user, selected from the crowd view (Fig.~\ref{fig:teaser}(E)). This helps clinical technicians compare a user's accuracy and performance based on demographics (T1). On top of the consensus map view, we also visualize which navigation direction (antegrade or retrograde) the video segments are from (Fig.~\ref{fig:consensusmap}(d)). Fig.~\ref{fig:consensusmap} illustrates each component of our consensus map.

In this view, clinical technicians can detect anomalous users and video segments (T2). We can detect patterns via a string matching algorithm to facilitate clinical technicians. More specifically, we use the Jaro Winkler distance~\cite{winkler:90} to calculate the similarity between a selected user's responses and the rest. In this prototype, we highlight all users with the same pattern as well as users with the top five similar matches. After detecting anomalous users and video segments, clinical technicians can mark these as anomalous users or anomalous video segments.
We provide a consensus slider to find an appropriate threshold percentage of users for obtaining a strong consensus (T3). The consensus is computed based on the ratio of the number of users who marked a video segment as a polyp video segment to the total number of users who viewed the video segment. When a clinical technician selects a threshold, we mark a video segment as a polyp video segment if a percentage number of users greater or equal to the specified threshold have marked it as polyp.
Since sensitivity and specificity are the gold standard measures used in medical applications, we also calculate sensitivity (SE) and specificity (SP) based on the specified threshold as follows:
\begin{equation}
	SE = \frac{TP}{(TP+FN)}, \qquad SP=\frac{TN}{(FP+TN)}
\end{equation}
where TP, FP, TN, and FN represent the number of true positive, false positive, true negative, and false negative video segments based on our threshold.
The calculated SE and SP are displayed when clinical technicians change the threshold (see top of Fig.~\ref{fig:consensusmap}).

\begin{figure}[htb!]
\centering
\begin{subfigure}[]{
\includegraphics[width=0.22\textwidth]{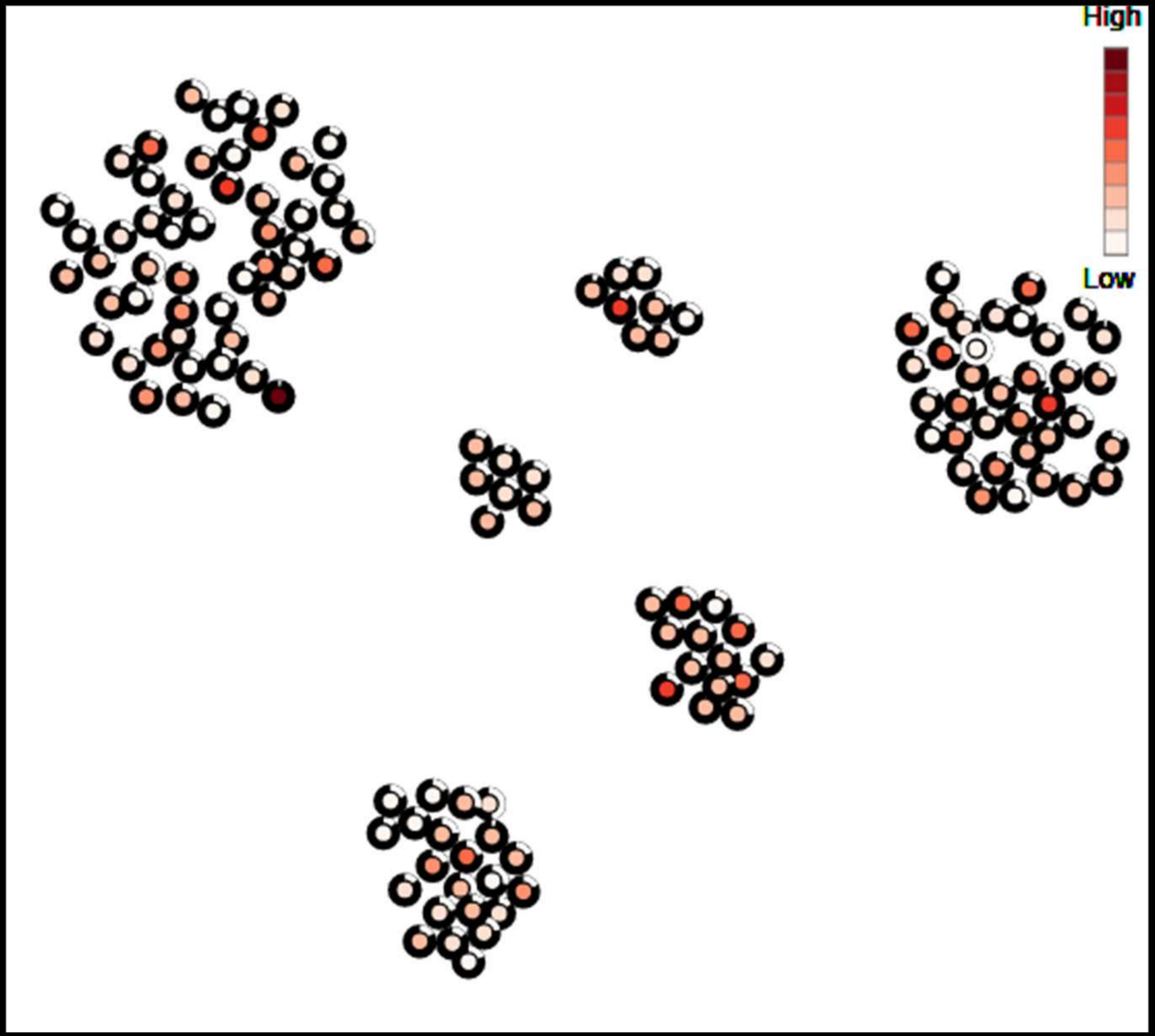}}
\end{subfigure}
\begin{subfigure}[]{
\includegraphics[width=0.22\textwidth]{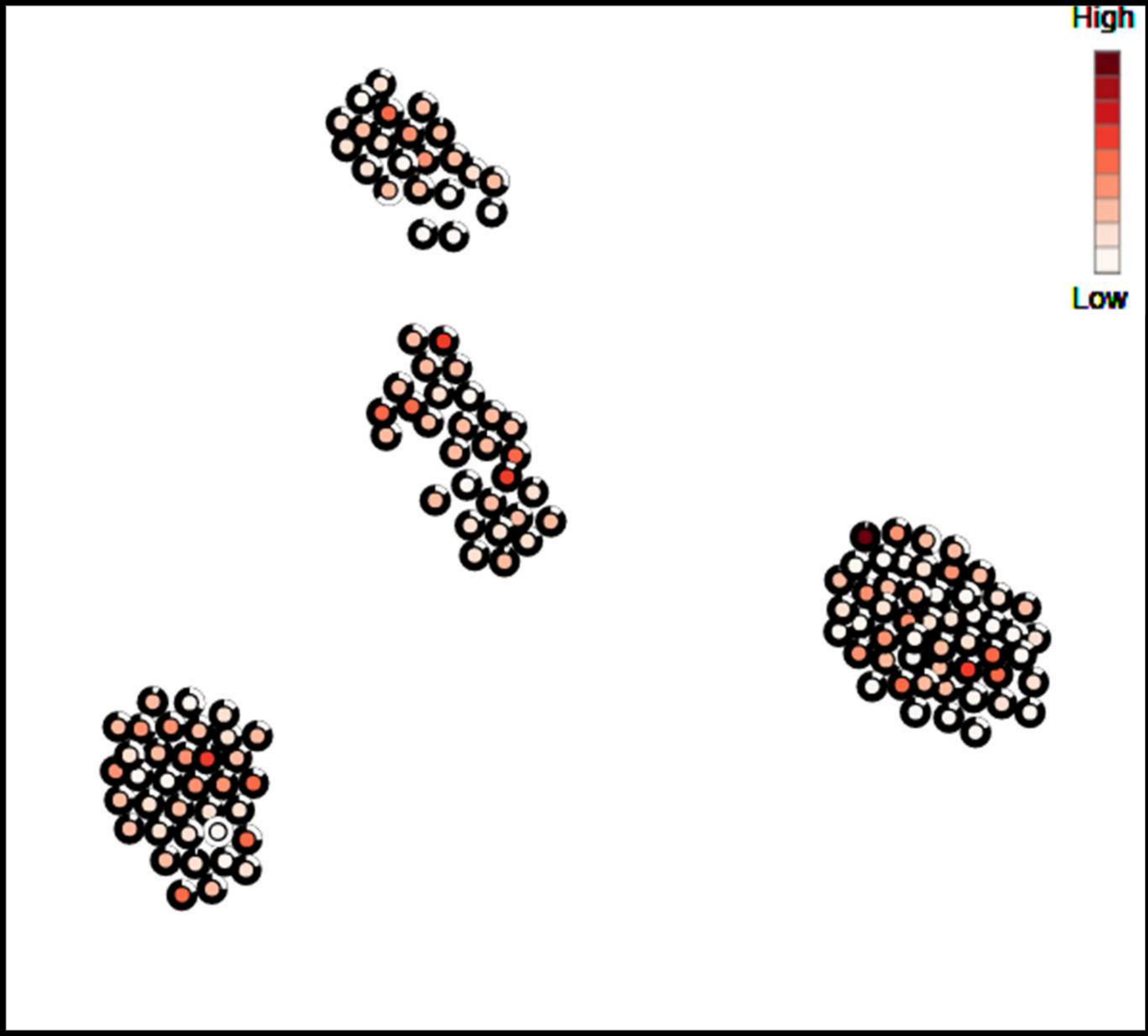}}
\end{subfigure}
\vspace{-4mm}
\caption{An example of our similarity view with predefined weights to cluster users based on the ``age'' parameter using (a) t-SNE, and (b) MDS.}
\label{fig:datareduction}
\vspace{-3mm}
\end{figure}

\subsection{Similarity View}
To A/B test for crowdsourcing platform specific parameters, we need to find similar users based on their demographics and rewards, and analyze their performance. Additionally, analysis of users' accuracy and timing on each video segment is necessary. We provide a similarity view to satisfy these requirements, which helps clinical technicians find optimal user performance and quality parameters (T1) and detect anomaly video segments (T2). Fig.~\ref{fig:similarityView}(b) and (c) illustrate our similarity views.
In the similarity view, each circle represents a user or a video segment, and we encode the number of true polyps detected or accuracy and the aggregated time to complete the task/video segment as the lightness of the circle (high:dark, low: light) and the length of the arc, respectively (Fig.~\ref{fig:similarityView}(a)).
We apply a force-directed algorithm to all circles in the similarity view to avoid overlap.
In order to compute the position of each user or video segment, we give the clinical technician the option to select well-known dimensionality reduction techniques: Multidimensional Scaling (MDS)~\cite{Kruskal:1964} or t-Distributed Stochastic Neighbor Embedding (t-SNE)~\cite{Maaten:2008}. These techniques can show similarity of users and colon video segments~\cite{Kwon:2016,Cao:2016}.
Both techniques have associated advantages and disadvantages. t-SNE has a non-convex optimization, and hence, the results might vary slightly across runs, while MDS has a convex optimization and gives more stable results across multiple runs~\cite{Maaten:2008}. On the other hand, MDS ignores local neighborhood information whereas t-SNE takes into account both global and local information, thus showing better clustering results~\cite{Ridgway:2012}.
Thus, we provide both techniques and let users make the selection. We encode the demographics and rewards information as categorical data and users' accuracy and timing on each video segment as numerical data. We use the overlap similarity metric~\cite{Stanfill:1986} for categorical data (e.g. analyzing a demographic parameter in Fig.~\ref{fig:datareduction}) and the Euclidean distance similarity metric for numerical data (e.g. analyzing video segments in Fig.~\ref{fig:similarityView} (c)) to compute MDS and t-SNE visualizations.
The overlap similarity metric and the Euclidean distance similarity metric are computed as follows:
\begin{eqnarray}
Overlap (A,B) = \frac{\sum_{i=1}^{N}[A_i \neq B_i] w_i}{N}\\[5pt]
Euclidean(A,B) = \sqrt{\sum_{i=1}^{N}(A_i-B_i)^2 w_i}
\end{eqnarray}
where $N$ is the number of dimensions or parameters, $A$ and $B$ represent two users or video segments, and $w_i$ is a weight for each parameter. All of the parameters are equally weighted.

In some cases, clinical technicians may want to cluster users and video segments based on specific parameters. For this purpose, we allow clinical technicians to select specific parameters with predefined weights to visualize this effect on the resultant clusters, as illustrated in Fig.~\ref{fig:datareduction}.

Our similarity view shows similarity between users based on their demographics and rewards. However, it cannot show details of demographics and rewards. In order to show details of selected users' information, we use the crowd view.

\begin{figure*}[tb]
\centering
\includegraphics[width=1.0\textwidth]{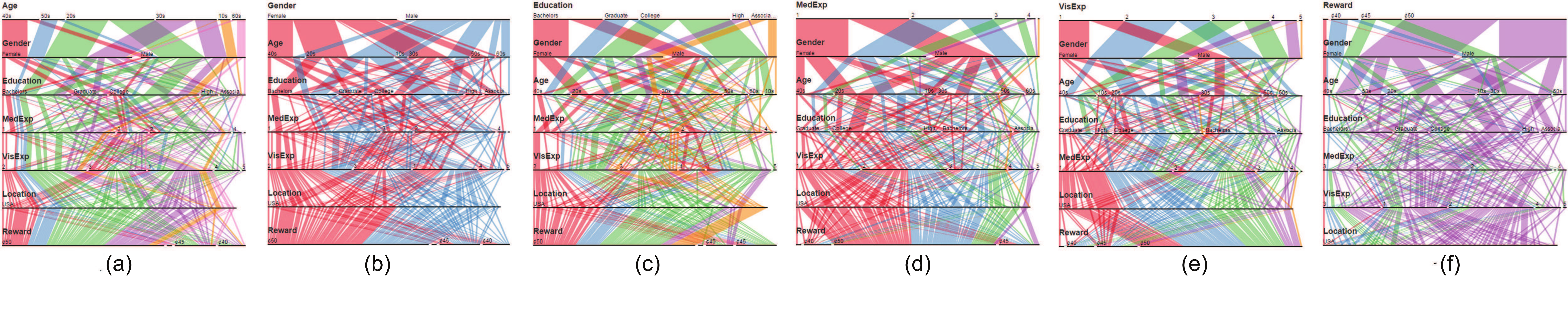}
\vspace{-10mm}
\caption{Variations of our crowd view by focusing on different parameters such as (a) age, (b) gender, (c) education level, (d) medical expertise, (e) visualization expertise, and (f) rewards.}
\label{fig:ps}
\vspace{-3mm}
\end{figure*}

\subsection{Crowd View}
As we mentioned earlier, our demographics and rewards data from each user is encoded as categorical data. In order to show the distributions as well as the relationships between these parameters (T1), we use the Parallel Sets technique~\cite{Kosara:2006}. This allows clinical technicians to visualize the overall distribution of each parameter by reordering them via drag and drop. We assign different colors to the elements in the parameter, at the top of the parallel set in the crowd view. All colors are selected from ColorBrewer~\cite{Harrower:2003}. Additionally, when clinical technicians select specific parameters for clustering in the similarity view, we automatically reorder our parallel sets and the selected parameters are moved to the top. This helps clinical technicians visualize how these selected parameters are distributed. Fig.~\ref{fig:ps} shows examples of our crowd view.

\subsection{Timeline Filtering View}
Our consensus map and similarity view focus on an individual user and/or a video segment.
Thus, these views cannot show all the information, for example, regarding how multiple users have performed on a complete colon dataset and their respective accuracies.
We need a broader view to manage datasets easily. For this purpose, we provide a timeline filtering view, as shown in Fig.~\ref{fig:teaser}(A), where a dataset created at a specific date is encoded as a bar. The height of a bar indicates the number of users who inspected the dataset and the color represents an average of their accuracy. This view not only provides an overview of all the datasets but can also be used to filter the datasets to focus on a specific dataset, which in turn is shown in the main views. Apart from allowing the user to narrow down to a specific dataset for which the date is known, we can use this view for a patient population level analysis for a given time period. This can be useful in figuring out the incidence of cancer rates in a particular time frame and perhaps help hospitals allocate appropriate resources for its treatment.

\begin{figure}[ht!]
\centering
\includegraphics[width=0.20\textwidth]{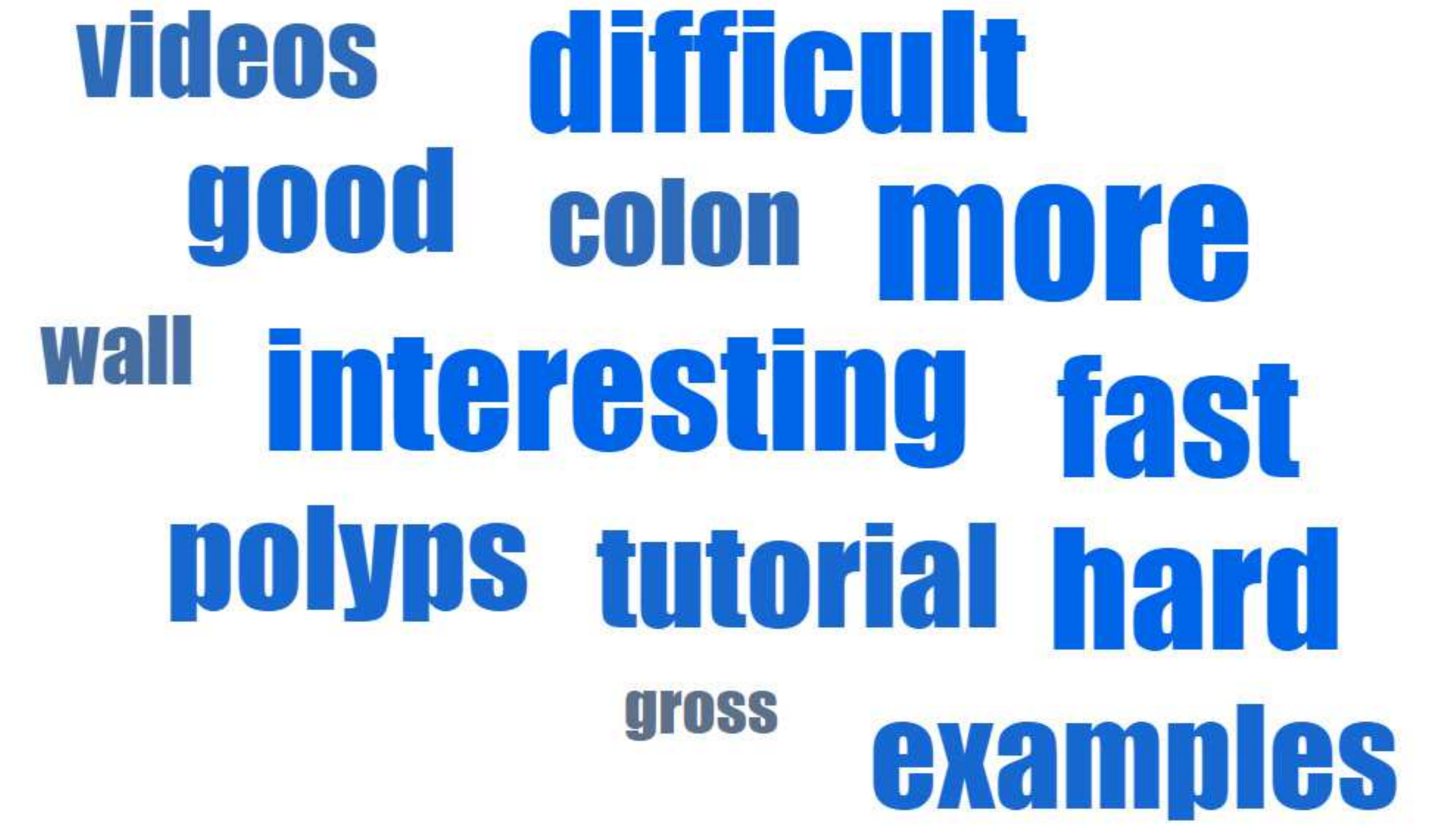}
\vspace{-3mm}
\caption{Our word cloud view for higher FOV and higher speed dataset.}
\label{fig:wordCloud}
\vspace{-2mm}
\end{figure}

\subsection{Aggregated Textual Information}
Fig.~\ref{fig:teaser}(B) shows an example of our aggregated textual information view. Under the Overview tab, this view provides basic aggregated textual information about the selected dataset in our timeline filtering view (Fig.~\ref{fig:teaser}(A)), including an average of all information described in Table~\ref{Tab:typeofdata}, and distributions for sensitivity and specificity. It helps clinical technicians to determine an appropriate threshold (T3) since varying the consensus rate from 0\% to 100\% might be tedious.
\begin{table}[h]
\renewcommand{\arraystretch}{1.1} 
\caption{Specificity and sensitivity for the crowd (based on the 50\% crowd consensus rate) and a medical expert in each case. \emph{Note the mismatch for case 1 due to the VC application parameter setting}.}
\label{Tab:result}
\vspace{-2mm}
\centering
\begin{tabular}{c|c|c|c}
  \hline
  Case	& User Type		& Specificity & Sensitivity 	 \\
  \hline   \hline
 \multirow{2}{*}{1} & crowd & 85.2\% & 65.2\% \\	\cline{2-4}
 & expert	& 72.7\% & 92.9\% \\	
  \hline
\multirow{2}{*}{2} & crowd& 80.0\% & 82.4\%   \\ \cline{2-4}
& expert	& 87.2\% & 86.7\% \\	
  \hline
\end{tabular}
\vspace{-3mm}
\end{table}
Under the Details tab, clinical technicians can visualize the accuracy distribution of the selected user for the whole tasks s/he participated in as a line graph. It also shows detailed information of a video when a video is selected.

\subsection{Word Cloud}
When we create videos and present them to the crowd, there might be defects or other issues in the videos or in the crowd platform, which can skew the consensus results.
Although we can detect most of these issues using our consensus map view, we also allow the users to explicitly provide feedback on such issues. This can help with the analysis of borderline cases. The crowd can leave comments at the end of their task to point out any issues in the video segments. In order to visualize these comments effectively, we opt for the Word Cloud technique, which can provide a keyword summary of these comments~\cite{Viegas:2009}. It helps clinical technicians detect anomaly video segments (T2), as shown in  Figs.~\ref{fig:teaser}(G) and~\ref{fig:wordCloud}.

%% file: sec-usecase.tex
\begin{figure*}[htb!]
\vspace{-2mm}
\centering
\begin{subfigure}[]{
\includegraphics[height=0.20\textheight]{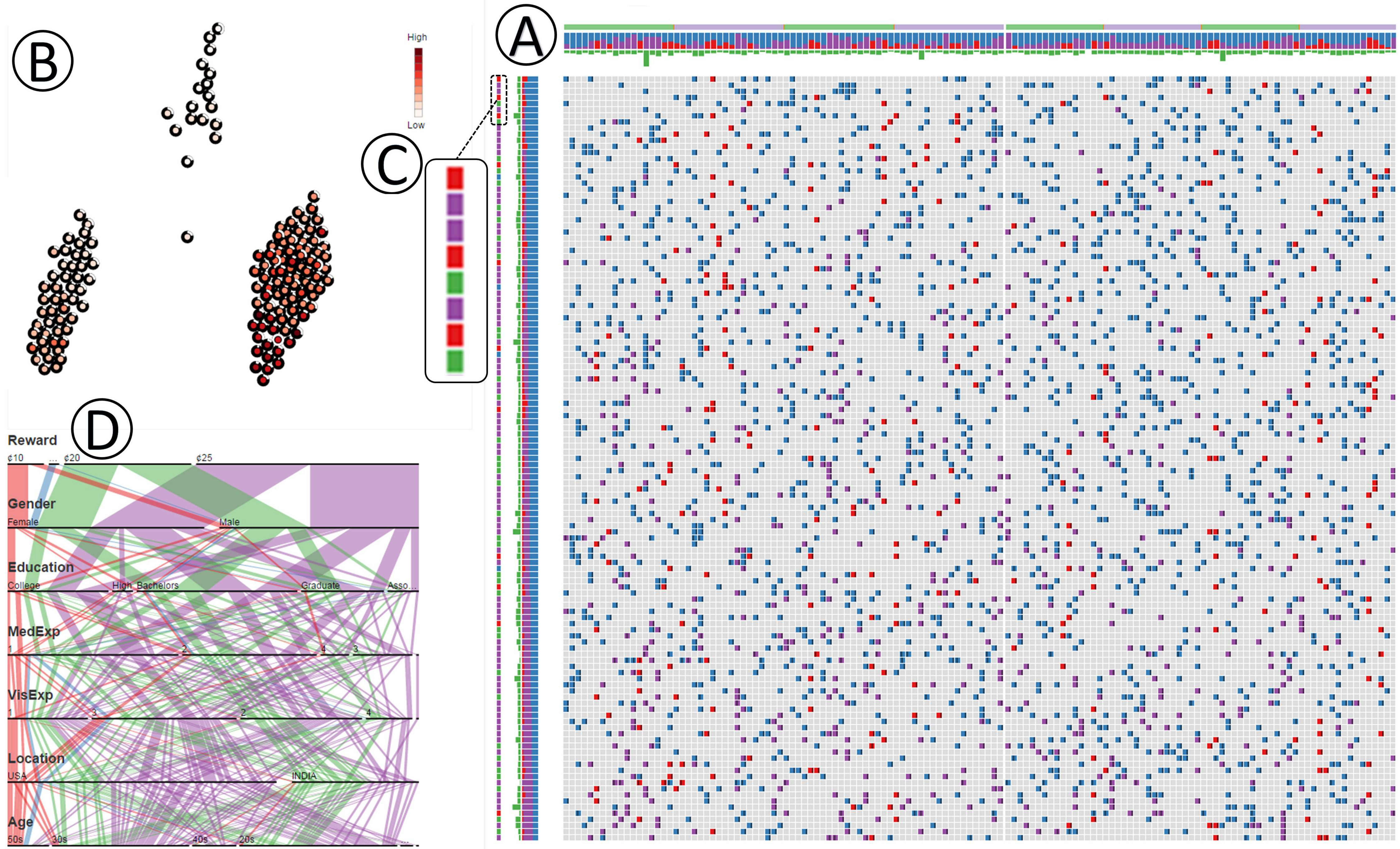}}
\end{subfigure}
\begin{subfigure}[]{
\includegraphics[height=0.20\textheight]{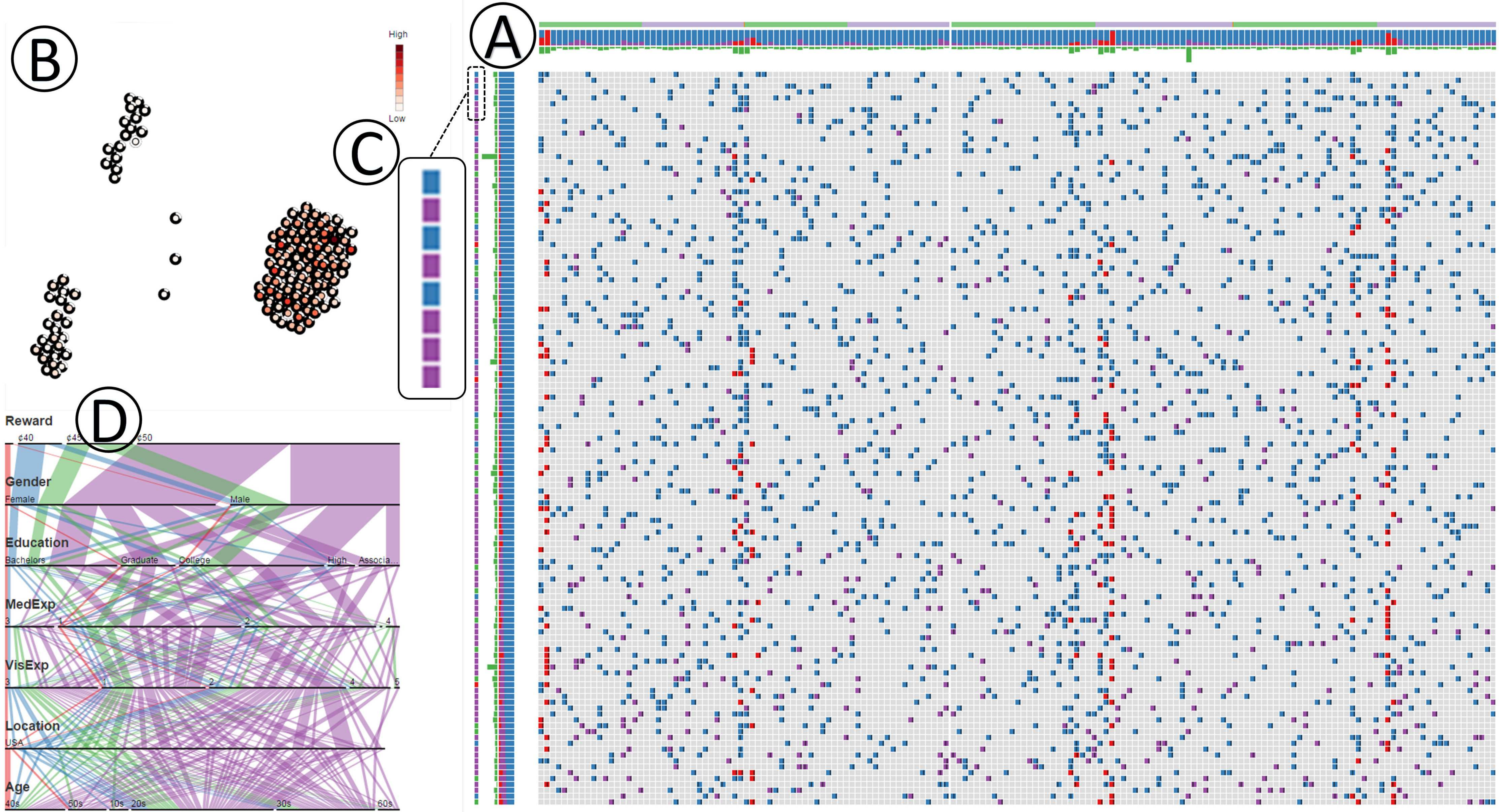}}
\end{subfigure}
\vspace{-4.5mm}
\caption{The effect of rewards in (a) higher FOV and speed dataset, and (b) lower FOV and speed dataset. In the consensus map view (A), users are sorted by accuracy and the demographics label (C) for each user shows its reward information, selected from the crowd view (D). In the similarity view (B), users are clustered based on the ``rewards'' parameter by assigning predefined weights to parameters. Both views show that the rewards have no effect on the user accuracy or performance in both the datasets.}
\label{fig:rewards}
\end{figure*}

\begin{figure*}[t]
\vspace{-4mm}
\centering
\begin{subfigure}[]{
\includegraphics[height=0.27\textheight]{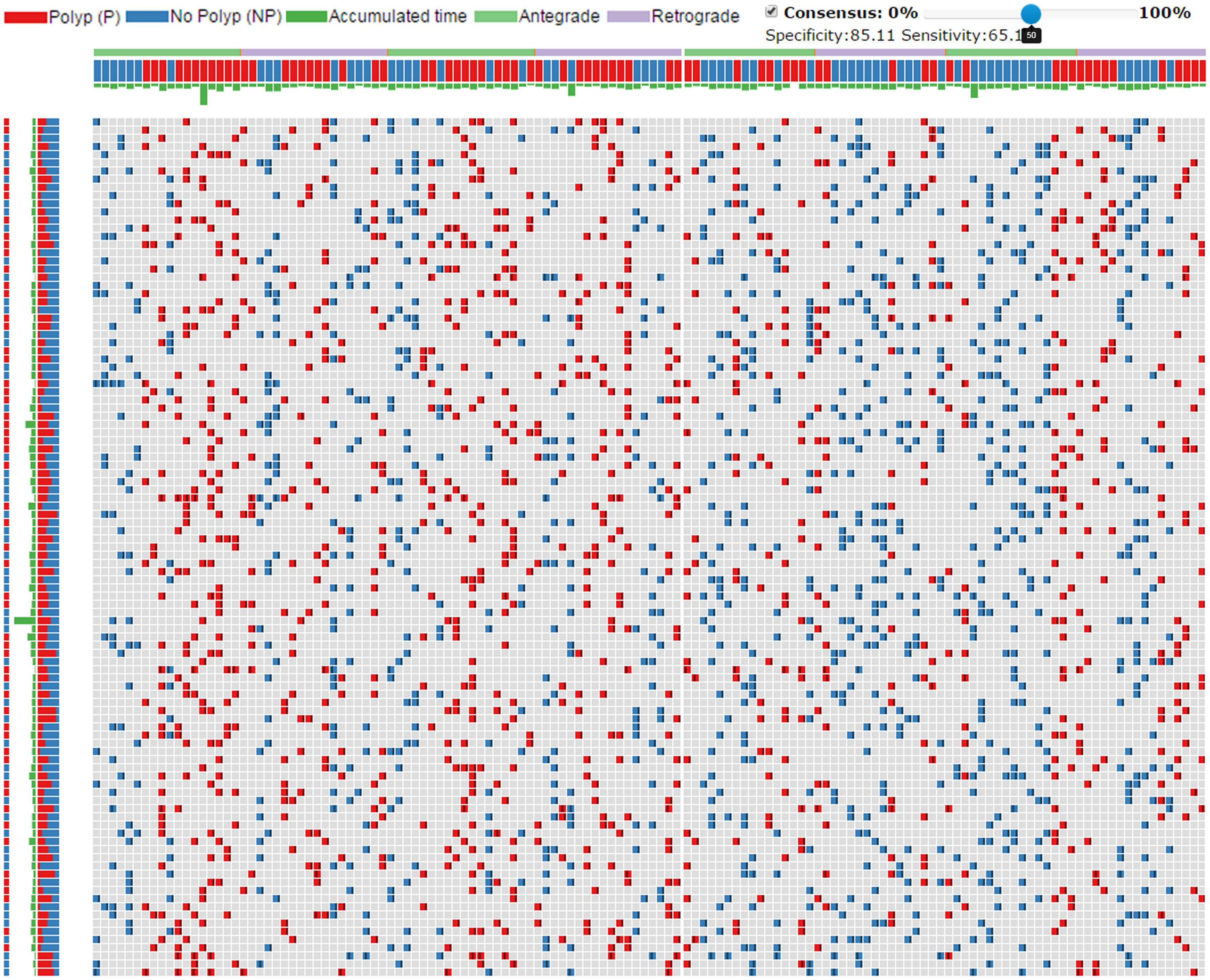}}
\end{subfigure}
\begin{subfigure}[]{
\includegraphics[height=0.27\textheight]{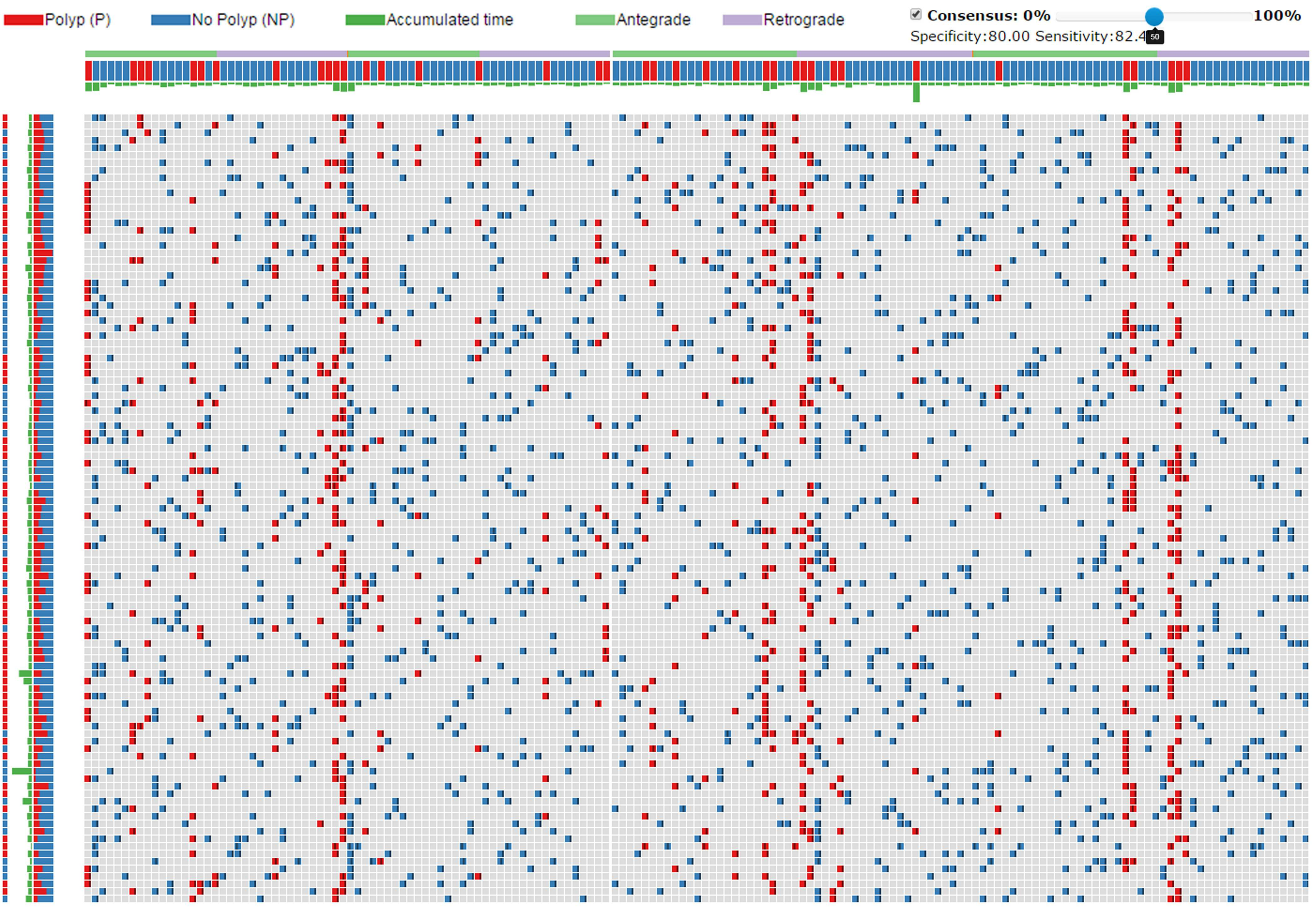}}
\end{subfigure}
\vspace{-4.5mm}
\caption{Our consensus map shows polyp and polyp-free video segments in (a) higher FOV and speed dataset, and (b) lower FOV and speed dataset based on the 50\% user consensus rate.}
\label{fig:consensus}
\vspace{-3mm}
\end{figure*}

\section{Evaluation}
\label{sec:casestudies}

We demonstrate the effectiveness of C$^2$A with VC datasets and crowd results. The VC datasets are real patient datasets from Stony Brook University Hospital that were properly anonymized before usage. To test different parameters of the video segment settings, we created two different datasets with the application specific parameters, such as the field-of-view and the speed of fly-through. The first dataset with 120 degrees field-of-view and 100 frames/second fly-through speed was presented to 136 non-expert users on MTurk and the second dataset with 90 degrees field-of-view and 50 frames/second was presented to 144 non-expert users on MTurk. Moreover, for the first dataset we also tried different locations for the non-expert users (USA and India) to test the hypothesis that users from one location are more reliable than the other.
We presented the crowd with two 10-second training videos. We highlighted the polyps in these videos to show what these precursor lesions look like. In the future, we will increase our training set and include more complex cases, such as flat polyps (less than 10mm in size).
The application-specific parameters had a considerable effect on the final results (Table~\ref{Tab:result}), as demonstrated in the following sections. We also interviewed five VC-trained radiologists for feedback on our C{$^2$}A platform to gain more insights into the final VC requirements. Moreover, for both the datasets we had an expert radiologist go through all the video segments to help us test our crowd sensitivity and specificity against an expert radiologist. With the lower FOV and speed dataset, we were able to reach sensitivity and specificity comparable to the radiologist as discussed below. The ground truth for both datasets was obtained from the pathology data obtained after performing an optical colonoscopy on the patient after polyp detection using the VC application.

\subsection{Case Studies}

\subsubsection{Case 1: Higher FOV and Higher Speed Dataset}
In the first dataset, we leveraged our analytics platform to explore and analyze two patient VC datasets with higher FOV and higher speed of navigation. We generated 136 video segments from these two patient VC datasets, which were each 10 seconds long. We also let the users fill out a survey at the end of every task to help identify any issues with the data or provide general feedback.

We found several issues in our video creation based on the user feedback (T2), as shown in Fig.~\ref{fig:wordCloud}. More specifically, we found several keywords, such as ``difficult'' and ``fast'', to highlight some issues. When these keywords were selected, C{$^2$}A highlighted the corresponding users and video segments in the linked views from which we were able to identify issues such as high distortion and fast speed of navigation. Due to the distortion incurred in the high field-of-view it was very difficult to distinguish between colon folds and polyps at times which led to a lack of consensus at the end.
Moreover, the speed of fly-through was too high for beginners, as shown from the word cloud in Fig.~\ref{fig:wordCloud}. 

We also tested for other demographics and incentive parameters to help correlate with user accuracy, reliability, and performance (T1). However, due to the severe distortions and resulting poor user accuracy in general, we were not able to find any clear patterns to test for hypotheses mentioned in previous crowdsourcing studies~\cite{Suri:2011,Heer:2010}, for example \emph{user location as a measure of reliability}, or \emph{the effect of rewards on the user accuracy or performance} (Fig.~\ref{fig:rewards}(a)).

Lastly, we explored our aggregated textual information view and consensus map to find an appropriate threshold of users to build a strong consensus (T3).
We viewed the bar charts under the overview tab to understand the distribution of specificity and sensitivity and found the best tradeoff between 45\% and 60\% user consensus. We then looked at our consensus view to understand these more closely (Fig.~\ref{fig:consensus}(a)). We changed the consensus rate between 45\% and 60\% and viewed the consensus view both with and without the ground truth to identify polyp-free video segments and false negative video segments.
In this dataset, we could not find an ideal consensus rate threshold because of poor user accuracy.



\subsubsection{Case 2: Lower FOV and Lower Speed Dataset}
In the second case study, we used C{$^2$}A to explore and analyze another two patient VC datasets with lower FOV and half the speed as compared to the previous dataset, based on the analytics performed with the previous dataset. In the second dataset, we generated 163 video segments from two patient VC datasets.

Similar to our first FOV dataset, we analyzed the aggregated task information in our consensus map to detect anomaly video segments and users (T2). In effect, we saw clear patterns emerging this time for polyp and polyp-free video segments based on the lower distortion and lower speed of fly-through, and for users who marked yes/no randomly. Based on these patterns, we performed pattern matching on our data and identified similar video segments and users.

More specifically, in order to detect anomaly users, we visualized datasets sorted by time-to-complete-the-task because some users randomly clicked answers without viewing video segments. As we expected, several users who spent little time to complete the task showed a pattern, for example one user clicked all ``Yes''. We marked them as anomaly users. We also wanted to know whether there were any users who spent more time than those anomaly users, but answered similarly. We picked anomaly users and found users with similar answers via our pattern matching method. There were several users who randomly clicked answers. We also marked them as anomaly users. We also compared these anomaly users against the ground truths or the majority responses for a set of video segments and found similar patterns. Similarly, for video segments where the distribution of the answers was random, we marked these video segments as anomalous. These segments would normally have lighting issues or some VC artifacts from preprocessing. Additionally, we also found an anomalous video segment in our similarity view, where only one video segment was clustered separately from other video segments. The majority of users identified it as a polyp video segment, but it was not. Due to lighting issues, a portion of a fold looked like a polyp.

Next, we clustered datasets based on each parameter and combination of parameters to observe the corresponding effect on user's performance (T1). We found that the rewards had no effect on the user accuracy or performance (Fig.~\ref{fig:rewards}(b)), as hypothesized in other crowdsourcing studies \cite{Heer:2010}. However, we found one interesting pattern, where users who claimed themselves as medical experts (4 and 5 level) had similar accuracy. Additionally, we selected the best and worst performing users, and the users with similar demographics and rewards in our similarity view. However, we did not find any similarity between them in terms of accuracy and timing.

Lastly, we explored our views to find an appropriate threshold of users as in the previous case (T3). We found that 50\% was the best threshold for our purpose (specificity: 80.0\%, sensitivity: 82.4\%). This threshold allowed almost 80\% of the video segments to be marked as polyp-free (Fig.~\ref{fig:consensus}(b)). This marking of the majority of the video segments as polyp-free has a huge potential for reducing the corresponding interpretation time for the medical experts by letting them focus on the video segments containing polyps.

\subsection{Interviews with Domain Experts}
We held regular discussions with VC-trained radiologists throughout the conception of this platform. For both FOV datasets, the radiologists were actively involved in marking all the video segments (generated from our pipeline) as polyp or polyp-free before presenting these videos to the crowd. Based on these medical expert annotations, we are able to compare the sensitivity and specificity of the medical experts with the non-expert users on MTurk. Moreover, this helps us in finding appropriate consensus thresholds to match the medical experts.

We also demonstrated the different interactive visual components of our platform to the experts.
The experts gave their feedback on our C{$^2$}A tool by putting up different hypotheses which we tested in the above case studies. In general, their feedback for C{$^2$}A was positive in terms of finding the best consensus to reach comparable sensitivity and specificity to the radiologists.
With brief explanation, they were able to appreciate the different views and how these views fit in the bigger picture. They especially liked the consensus view because it clearly showed the polyp and polyp-free regions based on the crowd consensus along with the individual user's performance. They also found the word cloud very interesting because it showed the keywords from the user feedback on the crowdsourced VC application. They found the crowd view a little confusing in the beginning when it displayed the distribution overview of each parameter. However, when a particular video or user segment data was selected, they were able to appreciate the linkage between the selected data and the parameters in the crowd view, which helped them understand the overall significance of the crowd view. Finally, the experts suggested extending our platform to incorporate lung nodule interpretation, which is a demanding procedure for the radiologists, yet conducive to crowdsourcing and hence ideal for C$^2$A.

%% file: sec-discussion.tex
\section{Discussion}
\label{sec:discussion}
The crowd statistics exhibit rich and valuable user information, which allow us to acquire a better understanding of user performance analysis and derive a set of considerable design principles.
Although in this paper we demonstrate the effectiveness of C{$^2$}A for VC, this platform can also be used for other crowdsourcing medical applications, such as virtual pancreatography (with four categories of precursor pancreatic cancer cysts), breast mammography (malignant and benign breast tissue), and lung nodules detection (nodules and nodule-free lung regions).

In our consensus map view, we are currently catering to two categories, namely polyp and polyp-free segments, but our approach can be extended to up to 12 distinguishable categories that can be useful in other general purpose crowdsouring applications. The 12 distinguishable categories are backed by previous research where users have been found to distinguish 12 different colors~\cite{ware:2004:IVP}.

In the following paragraphs, we discuss some of the limitations of our individual analytics components with respect to scalability, timing, word cloud generation, and variability in crowd statistics.

\emph{Scalability}: 
In the similarity view, we use non-overlapping techniques to avoid the overlap among data points and give a more accurate perspective with respect to individual data points. We can show up to two hundred users in this view without clutter. However, this can be a problem if the data points in our similarity view are increased beyond that threshold since the view can then become cluttered. We can resolve this issue by displaying time and the number of polyps/accuracy information separately and then applying Splatterplots~\cite{Mayorga:2013} to each one.

The consensus map currently requires a sliding bar to view several different colon datasets. This can hamper the aggregation analysis for these datasets and hence result in a strong reliance on the video segments view in the similarity view panel. We have, however, put different filtering mechanisms in place to deal with this issue, since our ultimate focus is the interpretation of an individual dataset at a time.
When clinical technicians select a dataset in the timeline filtering view, the consensus map shows all the details of individual segments in the selected dataset.
An alternative approach is that we can hierarchically group users or videos in the consensus view and the similarity view based on the similarity metric~\cite{Dang2015}, and collapse and expand these views as needed.


\emph{User and video segment timing}: We assume that the user won't take more than a certain amount of time per video segment and hence currently we normalize the time based on the maximum time point, to compare timing between users and video segments. Our approach works well when there is no outlier in the timing. However, if the user takes significantly more time than the other users, then our plots can become skewed towards that specific time point. This can affect both timing information on the consensus map as well as on the arcs on the data points in the similarity view. We can discard these skewed data points for now or mark them as anomalies and do the normalization with these points excluded.

\emph{Word cloud}: We compute the word cloud based on frequency and not uniqueness which can hide important keywords from the user comments. We will incorporate this uniqueness aspect in the future to deal with this limitation.

\emph{Uncertainty}: At the moment we display only the averages for the video segments, user accuracy, and timing, which help clinical technicians find anomalous users and video segments. However, this hides the variability that is incurred by the user performance per video segment, which might be needed. We can add error bars to highlight this uncertainty and give a more accurate picture in this regard. 

%% file: sec-conclusion.tex
\section{Conclusion and Future Work}
\label{sec:conc}

In this work, we presented a crowd consensus analytics platform for VC to build a consensus on polyp and polyp-free video segments, to detect crowd and video segment anomalies, and to improve the workers' quality based on A/B testing of platform and application-specific parameters. We showed that clinical technicians can use our analytics platform to improve crowd consensus for polyp and polyp-free video segments, which helps in achieving sensitivity and specificity comparable to expert radiologists. In effect, this can help discard a majority of the polyp-free segments and allows the radiologist to focus on the video segments marked as polyps and hence reduce the overall VC interpretation time per case. Based on our analytics platform, we have also highlighted specific design principles which are applicable to any other application with similar data characteristics and task requirements to the one introduced here.

In the future, we will leverage retrospective VC data with known ground truth to find non-expert users with high polyp detection accuracy. These users will be invited to interpret the prospective VC cases. We will then test the effectiveness of C$^2$A in a real clinical workflow. This will entail the following: (1) clinical technician runs the pre-processing step on the anonymized patient VC data and uploads the video segments to the crowdsourcing platform; (2) clinical technician retrieves the crowd responses from the crowdsourcing platform and visualizes these responses in C$^2$A; (3) clinical technician builds the crowd consensus on the individual video segments using C$^2$A; (4) C$^2$A incorporates the crowd consensus results into a commercial VC system, such as Viatronix V3D Colon (which already provides APIs for annotating different regions of the VC data); (5) the radiologist analyzes the corresponding VC patient data in the clinical system and makes the final diagnosis using the C$^2$A crowd consensus data which is either displayed as color-coded regions (based on the consensus) or incorporated in the navigation speed with the regions marked as polyp-free (with a high consensus) traversed at a higher speed than the regions containing polyps.